\documentclass[nohyperref]{article}

\usepackage{microtype}
\usepackage{graphicx}
\usepackage{subfigure}
\usepackage{booktabs} 

\usepackage{hyperref}

\usepackage[dynnworkshop]{icml2022}

\usepackage{amsmath}
\usepackage{amssymb}
\usepackage{mathtools}
\usepackage{amsthm}

\usepackage[capitalize,noabbrev]{cleveref}

\theoremstyle{plain}

\theoremstyle{definition}

\theoremstyle{remark}

\usepackage[textsize=tiny]{todonotes}

\icmltitlerunning{Fault-Tolerant Collaborative Inference through the Edge-PRUNE Framework}

\begin{document}

\twocolumn[
\icmltitle{Fault-Tolerant Collaborative Inference through \\
           the Edge-PRUNE Framework}

\icmlsetsymbol{equal}{*}

\begin{icmlauthorlist}
\icmlauthor{Jani Boutellier}{xxx}
\icmlauthor{Bo Tan}{yyy}
\icmlauthor{Jari Nurmi}{yyy}
\end{icmlauthorlist}

\icmlaffiliation{xxx}{School of Technology and Innovations, University of Vaasa, Vaasa, Finland}
\icmlaffiliation{yyy}{Faculty of Information Technology and Communication Sciences, Tampere University, Tampere, Finland}

\icmlcorrespondingauthor{Jani Boutellier}{jani.boutellier@uwasa.fi}

\icmlkeywords{Machine Learning, Collaborative Inference, Fault Tolerance}

\vskip 0.3in
]

\printAffiliationsAndNotice{} 

\begin{abstract}
Collaborative inference has received significant research interest in machine learning as a vehicle for distributing computation load, reducing latency, as well as addressing privacy preservation in communications. Recent collaborative inference frameworks have adopted dynamic inference methodologies such as early-exit and run-time partitioning of neural networks. However, as machine learning frameworks scale in the number of inference inputs, e.g., in surveillance applications, fault tolerance related to device failure needs to be considered. This paper presents the Edge-PRUNE distributed computing framework, built on a formally defined model of computation, which provides a flexible infrastructure for fault tolerant collaborative inference. The experimental section of this work shows results on achievable inference time savings by collaborative inference, presents fault tolerant system topologies and analyzes their cost in terms of execution time overhead.
\end{abstract}

\section{Introduction}
\label{sec:intro}
Since a few years already, the execution of machine learning workloads has been moving from servers and the cloud to less powerful platforms, such as embedded and mobile devices. A considerable hindrance to this progress has been the significant computation load of machine learning inference, especially related to deep neural network (DNN) architectures. In order to match neural network complexity with computation platform resources, several different approaches have been developed: lightweight architectures such as MobileNets \cite{howard2017mobilenets} attempt to maintain high inference accuracy despite drastically reduced number of trainable parameters; post-hoc optimizations such as dense layer pruning \cite{zhu2017prune}, separable convolutions \cite{jaderberg2014speeding} and weight quantization \cite{courbariaux2016binarized} reduce inference time by approximating the original trained neural architecture, whereas neural accelerators \cite{skillman2020technical, han2016eie} leverage specialized hardware to speed up inference.

Orthogonal to the aforementioned techniques, distributed and collaborative inference have emerged as a notable branch of research. In these approaches, the neural network inference workload is distributed between low-resource \textit{endpoint devices} and high performance \textit{edge servers} (or the cloud); early milestone works of this direction are Neurosurgeon \cite{kang2017neurosurgeon} and DDNN \cite{teerapittayanon2017distributed}. In conjunction with collaborative inference, several dynamic neural network techniques have been successfully adopted: for instance, early-exit \cite{teerapittayanon2017distributed} can terminate inference at intermediate layers saving on communication bandwidth, whereas dynamic onloading \cite{almeida2021dyno} decides at runtime the target execution platform of DNN layers.

\begin{figure}[t]
\vskip 0.2in
\begin{center}
\centerline{\includegraphics[width=0.85\columnwidth]{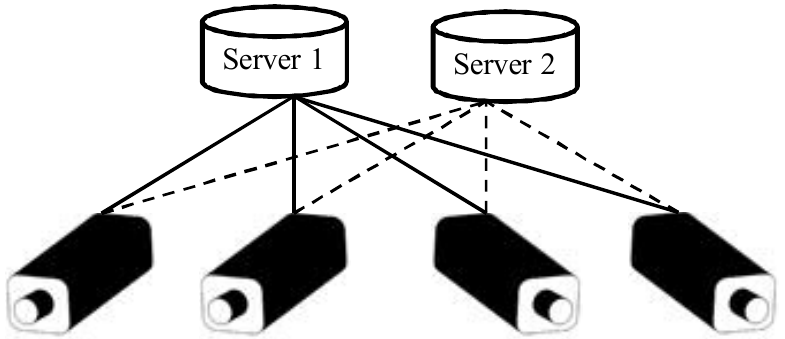}}
\caption{An example of a surveillance camera system with fault tolerance achieved by redundant nodes. The interconnection pattern equals to a $K_{2,4}$ complete bipartite graph, where 2 is the number of edge servers and 4 is the number of endpoint devices.}
\label{fig:overview}
\end{center}
\vskip -0.4in
\end{figure}

Collaborative inference can be used to target a variety of optimization objectives: reducing endpoint device computation load, end-to-end latency optimization, energy reduction or server workload reduction. A less frequently mentioned by-product of collaborative inference is privacy preservation. This topic has been extensively studied in \cite{he2019model}, where key observations point out that malicious \textit{model inversion attacks} against collaborative inference are significantly less effective if they cannot access the feature vectors produced by the early neural network layers -- especially if inference has passed one or more dense layers of the DNN. In terms of collaborative inference this means that the endpoint device should perform the inference of as many early DNN layers as possible before transmitting the intermediate feature vectors over a network interface to server processing.

This paper addresses the topic of collaborative inference, especially from the point of multiple inference inputs and system fault tolerance. An example scenario for multi-input collaborative inference is a smart surveillance camera system (see Figure~\ref{fig:overview}), where several smart cameras have been deployed across a site for performing object detection and/or tracking. For privacy preservation, the smart cameras perform the inference of early DNN layers, and transmit the intermediate feature vectors to a local edge server for completing the inference.

In safety critical areas, node (endpoint device or server) failures related to malicious actions or hardware faults needs to be taken in account. Concretely, the surveillance system should be able to continue its operation if one or more of the endpoint devices become disabled, or even in the more severe case of server failure. Figure~\ref{fig:overview} illustrates a highly redundant configuration, where a single failure of any kind of resource (endpoint device, server or connection) does not incapacitate the overall system. Such redundancy evidently comes with a price, which in this case is the redundant server and the related connectivity. 

This study, related to fault tolerant collaborative inference, is realized around the open source\footnote{Available at https://gitlab.com/jboutell/vprf/-/tree/edge-prune} \textit{Edge-PRUNE} framework \cite{boutellier2022edge}. Edge-PRUNE is based on a formally defined \textit{model of computation}, and provides the necessary theoretical infrastructure for specifying and designing collaborative inference between one or more endpoint devices and servers. Besides the theoretical basis, the Edge-PRUNE framework includes a self-sustained runtime engine that is hardware and training framework agnostic, providing a software environment for both endpoint devices and servers. Although not detailed in this work, Edge-PRUNE also has inherent support for conditional computing.


\section{Related Work}
\label{sec:related}

Significant early works on collaborative inference were DDNN \cite{teerapittayanon2017distributed} and Neurosurgeon \cite{kang2017neurosurgeon}. DDNN proposed distributing inference across endpoint, edge and cloud resources, also introducing early exits for reducing communication. Neurosurgeon, on the other hand, presented a scheduler for intelligently distributing neural network layers across endpoint and server resources. Edgent \cite{li2018edge} further developed Neurosurgeon's concept by introducing DNN right-sizing, joint optimization of early exits and DNN partitioning. Edgent later evolved into Boomerang \cite{zeng2019boomerang} inspired by the early exit mechanism of BranchyNet \cite{teerapittayanon2016branchynet}. IONN \cite{jeong2018ionn} also continued in the vein of Neurosurgeon, however based on the offloading concept: the endpoint device can upload DNN partitions to an edge server for optimizing mobile device energy consumption, among other optimization goals. Similar to Neurosurgeon, also IONN is based on Caffe \cite{jia2014caffe}. Recently, SplitNets \cite{dong2022splitnets} combined neural architecture search with multi-input partition point search.

JointDNN \cite{eshratifar2019jointdnn} introduced a directed acyclic graph (DAG) based model for DNN partitioning, optimizing for energy and latency. Besides partitioning, JointDNN also considers layer compression, similar to the preceding work JALAD \cite{li2018jalad}, and the recent \textit{supervised compression} work \cite{matsubara2022supervised}. A graph-based modeling approach is also adopted by DADS \cite{hu2019dynamic}, the industrial effort Auto-Split \cite{banitalebi2021auto} and $D^3$ \cite{zhang2021dynamic}, enabling capturing of branched DNN topologies as opposed to simpler chain-like DNN structures. Finally, SPINN \cite{laskaridis2020spinn} and DynO \cite{almeida2021dyno} contribute to dynamic DNN partitioning, which is useful under, e.g., varying wireless network conditions.

An orthogonal approach to endpoint-server computation partitioning is taken by \cite{mao2017mednn, mao2017modnn,  zhao2018deepthings, gao2021edgesp} that propose partitioning DNN inference across multiple endpoint devices.

\section{The Edge-PRUNE Framework}
\label{sec:framework}

The Edge-PRUNE framework used in this study significantly differs from the related works in the sense that it is based on a formal model of computation. The overall computation scheme of Edge-PRUNE is \textit{dataflow}, similar to that of TensorFlow \cite{abadi2016tensorflow}. However, Edge-PRUNE goes further in computation modeling, by formalizing concepts such as data packaging, data rates, triggering of computations, and necessary conditions for deadlock-free conditional computations.

\subsection{Model of Computation}

The Edge-PRUNE framework relies on the VR-PRUNE model of computation \cite{boutellier22vr-prune}. In VR-PRUNE, a neural network is described as a directed graph $G = (A,F)$, where the \textit{actors} $A$ represent vertices that perform computation, such as inference of a DNN layer. The links $F$ of graph $G$ represent first-in-first-out (FIFO) buffers that carry data between actors. For each link $f \in F$, data is quantized into fixed-size \textit{tokens} that in the neural network context can be understood as feature vectors between layers. A FIFO $f \in F$ is connected to an actor $a \in A$ through a \textit{port} $p_a$, such that $fifo(p_a) = f$.

An actor $a \in A$ can \textit{fire} (perform a computation) when it has a sufficient number of input tokens available. To specify the required number of tokens, for each input port $p_a$ of actor $a$, a non-negative integer-valued \textit{token rate} $atr(p_a)$ is defined; once each input port of $a$ has at least $atr(p_a)$ number of tokens in the associated FIFO buffer $fifo(p_a)$, actor $a$ becomes \textit{enabled}, that is, ready to fire. The exact moment in time when an enabled actor fires can depend, for instance, on availability of compute resources. As dataflow models in general, also VR-PRUNE is inherently \textit{concurrent}: individual actors can execute in parallel, independent of others \cite{Lee87}.

The VR-PRUNE model of computation balances between expressiveness and analyzability: while being expressive enough for allowing conditional computations, the model simultaneously provides means for analyzing graph consistency. The following subsection illustrates how conditional computation is expressed using VR-PRUNE concepts.

\subsection{Conditional Computation}

\begin{figure}[t]
\vskip 0.2in
\begin{center}
\centerline{\includegraphics[width=0.80\columnwidth]{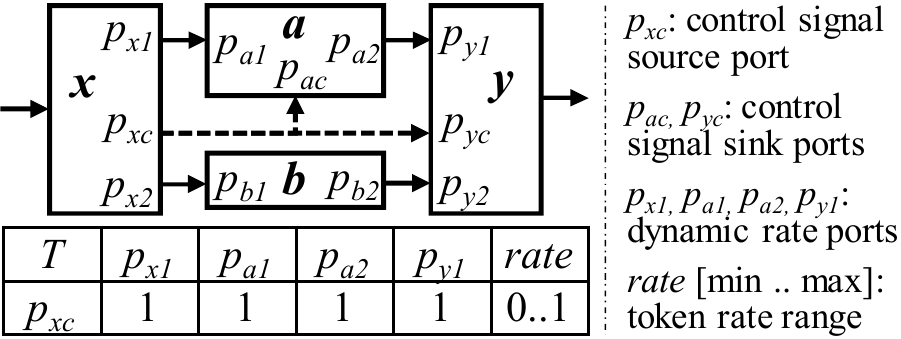}}
\caption{An Edge-PRUNE dynamic processing graph example.}
\label{fig:conditional}
\end{center}
\vskip -0.4in
\end{figure}

In order to maintain analyzability against graph deadlock and/or buffer overflow, the VR-PRUNE model restricts conditional computation to take place within \textit{dynamic processing (sub)graphs}, DPGs: within a DPG, conditional computation can be realized between two \textit{dynamic actors}. Figure~\ref{fig:conditional} shows a minimal case of a DPG: the dynamic actor $x$ provides a control signal (dashed connection) that at run time sets the input and output token rates of the dynamic actor $y$ and the \textit{dynamic processing actor} $a$. The range of allowed token rates and associated actor ports is expressed in the \textit{control table} $T$ of Figure~\ref{fig:conditional}: port $p_{xc}$ dynamically sets the token rate of ports $p_{x1}$, $p_{a1}$, $p_{a2}$ and $p_{y1}$ to either 0 or to 1. With token rates set to 1, $a$ becomes enabled, whereas token rate 0 disables execution of $a$. Actor $b$ does not receive such a control signal and thus maintains static token rates. 

\subsection{Distributed Computing and Fault Tolerance}
The Edge-PRUNE framework enables concurrent execution of actors both within a computing platform, and between computing platforms using a \textit{mapping specification}. Within a platform, each actor $a \in A$ is mapped for execution to a specific CPU core, or to the local GPU. The same mapping specification is also used to set the execution platform (endpoint device or server) of each actor. To this extent, Edge-PRUNE provides mapping exploration functionality, which auto-generates endpoint-server mapping alternatives for discovering the best DNN partition point for collaborative inference.

The Edge-PRUNE mapping specification allows generalizing distributed computing also to system configurations that consist of one or more endpoint devices and/or servers, e.g., for achieving fault tolerance. From the dataflow model viewpoint, computing node malfunction means that either a subgraph of $G$ ceases to produce tokens (endpoint failure), or a subgraph of $G$ stops consuming tokens (server failure). Maintaining operation for the remaining system requires that the dataflow application needs to overcome such unexpected subgraph failures without ending up in deadlock. Currently, Edge-PRUNE implements fault tolerance on the dataflow graph level, whereas extending the fault tolerance behavior to model-based graph consistency analysis remains a prospective for future work.

\subsection{Inference Engine}

The Edge-PRUNE computing functionalities have been implemented in the C language to a lightweight runtime library for Linux-based platforms, enabling deployment to embedded devices as well as to servers. This inference engine is independent of neural network training frameworks (e.g., TensorFlow), but allows leveraging DNN acceleration libraries such as Intel oneDNN or ARM CL. Edge-PRUNE has deeply inbuilt support for GPU leverage, but can equally well operate on GPU-less platforms. Distributed computing functionality has been implemented using Linux Sockets, such that the endpoint devices are expected to establish an \texttt{ssh} connection to the server(s), delegating data security issues to the level of \texttt{ssh} connections. Fault tolerance functionality is implemented on the actor port level: ports responsible for inter-platform communication monitor and adjust to remote platform liveness by Linux socket error conditions: broken pipe, connection reset, no data sent. Finally, Edge-PRUNE does not constrain the nature (wired or cable) or number (shared or point-to-point) of connections between endpoints and servers.

\section{Experiments}
\label{sec:experiments}

\begin{figure}[t]
\vskip 0.2in
\begin{center}
\centerline{\includegraphics[width=\columnwidth]{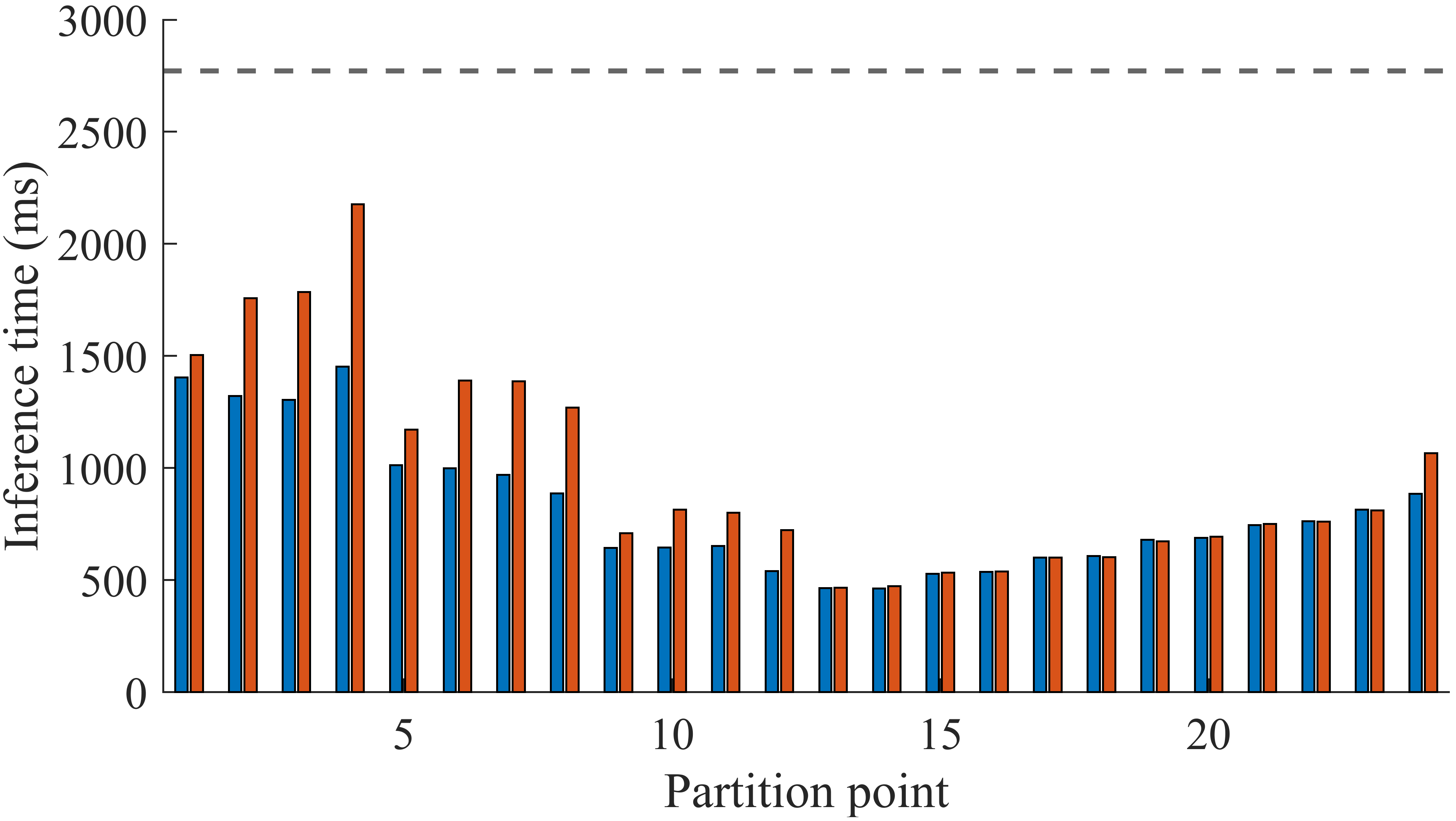}}
\caption{Collaborative inference partition point exploration for SSD-MobileNet v1 (300$\times$300 input) under Edge-PRUNE. Bars reflect endpoint inference + communication time (blue: 100 Mbit Ethernet, red: 16 Mbit WiFi); horizontal dashed line is endpoint-only inference time (all per 1 frame).}
\label{fig:mobilenet}
\end{center}
\vskip -0.2in
\end{figure}

\subsection{Collaborative Inference Partition Point Exploration}

Figure~\ref{fig:mobilenet} shows endpoint device inference time for the SSD-MobileNet v1 \cite{howard2017mobilenets, liu2016ssd} object detection CNN in a collaborative inference scenario where an ODROID N2 single-board  computer (Hexacore ARM+GPU) acts as the endpoint device, and an Intel Core i7-8650U based platform as the edge server, interconnected by 100 Mbit Ethernet or 16 Mbit WiFi. The full-precision CNN (32 bit float) is implemented as an Edge-PRUNE application such that each Conv-BNorm-ReLU layer triplet is wrapped inside a dedicated actor, forming a graph of 53 actors and 69 links. Endpoint device inference time was explored by shifting the endpoint/server partition point actor-by-actor from inference input towards inference output, resulting in Figure~\ref{fig:mobilenet}. The graph shows that partition points 13 and 14 provide minimal endpoint inference time, 6.0$\times$ higher than endpoint-only inference. On the endpoint device, the CNN layers inside Edge-PRUNE actors were implemented using ARM Compute Library functions.

\subsection{Inference Time and Fault Tolerance}
\label{ssec:tolerance}
Both virtual and physically distributed system configurations were used validate Edge-PRUNE collaborative inference and system behavior on computing node failures.

\textbf{Virtual environment.} \cref{throughput-table} illustrates performance scaling of Edge-PRUNE collaborative inference for single and dual-server configurations with $1 \le n \le 6$ endpoint devices and complete bipartite graph $K_{m,n}$ interconnection topology. Per-frame processing time was measured in a virtual distributed environment: each endpoint process and each server process was assigned to a dedicated core on the 8-core Intel Core i7-8650U processor, and connections between endpoints and servers were handled over the Linux \textit{loopback} network interface, which allowed using exactly the same server and endpoint software configurations as in a physically distributed system. Each endpoint performed the inference of 7 initial layers (Conv2D-ReLU-MaxPool-Conv2D-ReLU-MaxPool-Dense) of a vehicle classification CNN \cite{xie2016resource}, whereas each server process performed the inference of the last 5 layers of the same CNN. The 7 CNN layers of each endpoint subgraph were realized into 4 actors, whereas the 5 layers of the server subgraph were wrapped inside a single actor; therefore the largest $K_{2,6}$ configuration consisted of 26 actors and 30 FIFOs. On the endpoint side, Conv2D layer inference was implemented by the Intel oneDNN library. \cref{throughput-table} shows: a) adding a second server for fault tolerance causes insignificant processing time overhead, and b) adding an endpoint increases per-node inference time by 11\% on average. Endpoint and server fault behavior was observed by abrupt termination of endpoint/server processes, showing that the remaining system continues collaborative inference as expected. 

\textbf{Heterogeneous distributed system.} To confirm Edge-PRUNE collaborative inference functionality and behavior on node fault on a physically distributed system, $K_{1,2}$ and $K_{2,1}$ configurations were established using a heterogeneous set of nodes: an Intel Core i7-8650U workstation, an ODROID N2, and an Intel Atom N270 based platform, all running Ubuntu Linux, and interconnected over 100 Mbit Ethernet. Using the same vehicle classification CNN \cite{xie2016resource}, the heterogeneous configuration was used to validate that a) Edge-PRUNE recovers from communication breaks (temporarily disconnected cable), and b) maintains operation on permanent node failure (server power-down for $K_{2,1}$ and endpoint power-down for $K_{1,2}$).

\begin{table}[t]
\caption{Vehicle image classification collaborative inference time per frame in milliseconds for $1 \le m \le 2$ servers and $1 \le n \le 6$ endpoint devices. Upper half: endpoint, lower half: server.}
\label{throughput-table}
\vskip 0.15in
\begin{center}
\begin{small}
\begin{sc}
\begin{tabular}{lcccccr}
\toprule
N. of endpoints & 1 & 2 & 3 & 4 & 5 & 6 \\
\midrule
Single-server & 4.7 & 4.9 & 5.2 & 5.4 & 6.2 & 7.0 \\
Dual-server & 4.9 & 5.1 & 5.2 & 5.5 & 6.3 & 7.1 \\
\\
Single-server & 4.8 & 5.7 & 6.5 & 7.1 & 8.7 & 9.7 \\
Dual-server & 4.9 & 5.8 & 6.5 & 7.1 & 8.9 & 9.6 \\
\bottomrule
\end{tabular}
\end{sc}
\end{small}
\end{center}
\vskip -0.1in
\end{table}

\section{Conclusion}
\label{sec:conclusion}

In this work the topic of collaborative inference fault tolerance was studied for configurations consisting of one or more network-connected edge servers and one or more endpoint devices. The experimental study was done using the Edge-PRUNE framework, which was shown to scale successfully with an increasing number of endpoint devices and/or edge servers, and furthermore was shown to be capable of continuing operation after computing node failure.


\begin{thebibliography}{33}
\providecommand{\natexlab}[1]{#1}
\providecommand{\url}[1]{\texttt{#1}}
\expandafter\ifx\csname urlstyle\endcsname\relax
  \providecommand{\doi}[1]{doi: #1}\else
  \providecommand{\doi}{doi: \begingroup \urlstyle{rm}\Url}\fi

\bibitem[Abadi et~al.(2016)Abadi, Barham, Chen, Chen, Davis, Dean, Devin,
  Ghemawat, Irving, Isard, et~al.]{abadi2016tensorflow}
Abadi, M., Barham, P., Chen, J., Chen, Z., Davis, A., Dean, J., Devin, M.,
  Ghemawat, S., Irving, G., Isard, M., et~al.
\newblock Tensorflow: A system for large-scale machine learning.
\newblock In \emph{{USENIX} Symposium on Operating Systems Design and
  Implementation ({OSDI})}, pp.\  265--283, 2016.

\bibitem[Almeida et~al.(2021)Almeida, Laskaridis, Venieris, Leontiadis, and
  Lane]{almeida2021dyno}
Almeida, M., Laskaridis, S., Venieris, S.~I., Leontiadis, I., and Lane, N.~D.
\newblock {DynO}: Dynamic onloading of deep neural networks from cloud to
  device.
\newblock \emph{ACM Transactions on Embedded Computing Systems (TECS)}, 2021.

\bibitem[Banitalebi-Dehkordi et~al.(2021)Banitalebi-Dehkordi, Vedula, Pei, Xia,
  Wang, and Zhang]{banitalebi2021auto}
Banitalebi-Dehkordi, A., Vedula, N., Pei, J., Xia, F., Wang, L., and Zhang, Y.
\newblock Auto-split: a general framework of collaborative edge-cloud ai.
\newblock In \emph{ACM SIGKDD Conference on Knowledge Discovery \& Data
  Mining}, 2021.

\bibitem[Boutellier et~al.(2022{\natexlab{a}})Boutellier, Ma, Wu, Khan, and
  Bhattacharyya]{boutellier22vr-prune}
Boutellier, J., Ma, Y., Wu, J., Khan, M., and Bhattacharyya, S.~S.
\newblock {VR-PRUNE}: Decidable variable-rate dataflow for signal processing
  systems.
\newblock \emph{IEEE Transactions on Signal Processing}, 70:\penalty0 1819 --
  1833, 2022{\natexlab{a}}.

\bibitem[Boutellier et~al.(2022{\natexlab{b}})Boutellier, Tan, and
  Nurmi]{boutellier2022edge}
Boutellier, J., Tan, B., and Nurmi, J.
\newblock Edge-prune: Flexible distributed deep learning inference.
\newblock \emph{arXiv preprint arXiv:2204.12947}, 2022{\natexlab{b}}.

\bibitem[Courbariaux et~al.(2016)Courbariaux, Hubara, Soudry, El-Yaniv, and
  Bengio]{courbariaux2016binarized}
Courbariaux, M., Hubara, I., Soudry, D., El-Yaniv, R., and Bengio, Y.
\newblock Binarized neural networks: Training deep neural networks with weights
  and activations constrained to +1 or -1.
\newblock \emph{arXiv preprint arXiv:1602.02830}, 2016.

\bibitem[Dong et~al.(2022)Dong, De~Salvo, Li, Liu, Qu, Kung, and
  Li]{dong2022splitnets}
Dong, X., De~Salvo, B., Li, M., Liu, C., Qu, Z., Kung, H., and Li, Z.
\newblock Splitnets: Designing neural architectures for efficient distributed
  computing on head-mounted systems.
\newblock In \emph{Proceedings of the IEEE/CVF Conference on Computer Vision
  and Pattern Recognition}, pp.\  12559--12569, 2022.

\bibitem[Eshratifar et~al.(2019)Eshratifar, Abrishami, and
  Pedram]{eshratifar2019jointdnn}
Eshratifar, A.~E., Abrishami, M.~S., and Pedram, M.
\newblock {JointDNN}: an efficient training and inference engine for
  intelligent mobile cloud computing services.
\newblock \emph{IEEE Transactions on Mobile Computing}, 20\penalty0
  (2):\penalty0 565--576, 2019.

\bibitem[Gao et~al.(2021)Gao, Sun, Zhang, Mo, and Zhao]{gao2021edgesp}
Gao, Z., Sun, S., Zhang, Y., Mo, Z., and Zhao, C.
\newblock {EdgeSP}: Scalable multi-device parallel {DNN} inference on
  heterogeneous edge clusters.
\newblock In \emph{International Conference on Algorithms and Architectures for
  Parallel Processing}, pp.\  317--333. Springer, 2021.

\bibitem[Han et~al.(2016)Han, Liu, Mao, Pu, Pedram, Horowitz, and
  Dally]{han2016eie}
Han, S., Liu, X., Mao, H., Pu, J., Pedram, A., Horowitz, M.~A., and Dally,
  W.~J.
\newblock {EIE}: efficient inference engine on compressed deep neural network.
\newblock In \emph{ACM/IEEE International Symposium on Computer Architecture
  (ISCA)}, pp.\  243--254. IEEE, 2016.

\bibitem[He et~al.(2019)He, Zhang, and Lee]{he2019model}
He, Z., Zhang, T., and Lee, R.~B.
\newblock Model inversion attacks against collaborative inference.
\newblock In \emph{Annual Computer Security Applications Conference}, 2019.

\bibitem[Howard et~al.(2017)Howard, Zhu, Chen, Kalenichenko, Wang, Weyand,
  Andreetto, and Adam]{howard2017mobilenets}
Howard, A.~G., Zhu, M., Chen, B., Kalenichenko, D., Wang, W., Weyand, T.,
  Andreetto, M., and Adam, H.
\newblock {MobileNets}: Efficient convolutional neural networks for mobile
  vision applications.
\newblock \emph{arXiv preprint arXiv:1704.04861}, 2017.

\bibitem[Hu et~al.(2019)Hu, Bao, Wang, and Liu]{hu2019dynamic}
Hu, C., Bao, W., Wang, D., and Liu, F.
\newblock Dynamic adaptive {DNN} surgery for inference acceleration on the
  edge.
\newblock In \emph{IEEE Conference on Computer Communications}, 2019.

\bibitem[Jaderberg et~al.(2014)Jaderberg, Vedaldi, and
  Zisserman]{jaderberg2014speeding}
Jaderberg, M., Vedaldi, A., and Zisserman, A.
\newblock Speeding up convolutional neural networks with low rank expansions.
\newblock In \emph{British Machine Vision Conference}, 2014.

\bibitem[Jeong et~al.(2018)Jeong, Lee, Shin, and Moon]{jeong2018ionn}
Jeong, H.-J., Lee, H.-J., Shin, C.~H., and Moon, S.-M.
\newblock {IONN}: Incremental offloading of neural network computations from
  mobile devices to edge servers.
\newblock In \emph{ACM Symposium on Cloud Computing}, 2018.

\bibitem[Jia et~al.(2014)Jia, Shelhamer, Donahue, Karayev, Long, Girshick,
  Guadarrama, and Darrell]{jia2014caffe}
Jia, Y., Shelhamer, E., Donahue, J., Karayev, S., Long, J., Girshick, R.,
  Guadarrama, S., and Darrell, T.
\newblock Caffe: Convolutional architecture for fast feature embedding.
\newblock In \emph{ACM international conference on Multimedia}, 2014.

\bibitem[Kang et~al.(2017)Kang, Hauswald, Gao, Rovinski, Mudge, Mars, and
  Tang]{kang2017neurosurgeon}
Kang, Y., Hauswald, J., Gao, C., Rovinski, A., Mudge, T., Mars, J., and Tang,
  L.
\newblock Neurosurgeon: Collaborative intelligence between the cloud and mobile
  edge.
\newblock \emph{ACM SIGARCH Computer Architecture News}, 45\penalty0
  (1):\penalty0 615--629, 2017.

\bibitem[Laskaridis et~al.(2020)Laskaridis, Venieris, Almeida, Leontiadis, and
  Lane]{laskaridis2020spinn}
Laskaridis, S., Venieris, S.~I., Almeida, M., Leontiadis, I., and Lane, N.~D.
\newblock {SPINN}: synergistic progressive inference of neural networks over
  device and cloud.
\newblock In \emph{Annual International Conference on Mobile Computing and
  Networking}, 2020.

\bibitem[Lee \& Messerschmitt(1987)Lee and Messerschmitt]{Lee87}
Lee, E.~A. and Messerschmitt, D.~G.
\newblock Synchronous data flow.
\newblock \emph{Proceedings of the IEEE}, 75\penalty0 (9):\penalty0 1235--1245,
  1987.
\newblock ISSN 0018-9219.

\bibitem[Li et~al.(2018{\natexlab{a}})Li, Zhou, and Chen]{li2018edge}
Li, E., Zhou, Z., and Chen, X.
\newblock Edge intelligence: On-demand deep learning model co-inference with
  device-edge synergy.
\newblock In \emph{Workshop on Mobile Edge Communications}, 2018{\natexlab{a}}.

\bibitem[Li et~al.(2018{\natexlab{b}})Li, Hu, Jiang, Wang, Wen, and
  Zhu]{li2018jalad}
Li, H., Hu, C., Jiang, J., Wang, Z., Wen, Y., and Zhu, W.
\newblock {JALAD}: Joint accuracy-and latency-aware deep structure decoupling
  for edge-cloud execution.
\newblock In \emph{IEEE International conference on parallel and distributed
  systems (ICPADS)}, 2018{\natexlab{b}}.

\bibitem[Liu et~al.(2016)Liu, Anguelov, Erhan, Szegedy, Reed, Fu, and
  Berg]{liu2016ssd}
Liu, W., Anguelov, D., Erhan, D., Szegedy, C., Reed, S., Fu, C.-Y., and Berg,
  A.~C.
\newblock {SSD}: Single shot multibox detector.
\newblock In \emph{European Conference on Computer Vision}, 2016.

\bibitem[Mao et~al.(2017{\natexlab{a}})Mao, Chen, Nixon, Krieger, and
  Chen]{mao2017modnn}
Mao, J., Chen, X., Nixon, K.~W., Krieger, C., and Chen, Y.
\newblock {MoDNN}: Local distributed mobile computing system for deep neural
  network.
\newblock In \emph{Design, Automation \& Test in Europe Conference \&
  Exhibition (DATE)}, 2017{\natexlab{a}}.

\bibitem[Mao et~al.(2017{\natexlab{b}})Mao, Yang, Wen, Wu, Song, Nixon, Chen,
  Li, and Chen]{mao2017mednn}
Mao, J., Yang, Z., Wen, W., Wu, C., Song, L., Nixon, K.~W., Chen, X., Li, H.,
  and Chen, Y.
\newblock {MeDNN}: A distributed mobile system with enhanced partition and
  deployment for large-scale {DNNs}.
\newblock In \emph{IEEE/ACM International Conference on Computer-Aided Design
  (ICCAD)}, 2017{\natexlab{b}}.

\bibitem[Matsubara et~al.(2022)Matsubara, Yang, Levorato, and
  Mandt]{matsubara2022supervised}
Matsubara, Y., Yang, R., Levorato, M., and Mandt, S.
\newblock Supervised compression for resource-constrained edge computing
  systems.
\newblock In \emph{Proceedings of the IEEE/CVF Winter Conference on
  Applications of Computer Vision}, pp.\  2685--2695, 2022.

\bibitem[Skillman \& Eds{\"o}(2020)Skillman and
  Eds{\"o}]{skillman2020technical}
Skillman, A. and Eds{\"o}, T.
\newblock A technical overview of {Cortex-M55} and {Ethos-U55}: {Arm’s} most
  capable processors for endpoint {AI}.
\newblock In \emph{IEEE Hot Chips Symposium (HCS)}, 2020.

\bibitem[Teerapittayanon et~al.(2016)Teerapittayanon, McDanel, and
  Kung]{teerapittayanon2016branchynet}
Teerapittayanon, S., McDanel, B., and Kung, H.-T.
\newblock Branchynet: Fast inference via early exiting from deep neural
  networks.
\newblock In \emph{2016 23rd International Conference on Pattern Recognition
  (ICPR)}, pp.\  2464--2469. IEEE, 2016.

\bibitem[Teerapittayanon et~al.(2017)Teerapittayanon, McDanel, and
  Kung]{teerapittayanon2017distributed}
Teerapittayanon, S., McDanel, B., and Kung, H.-T.
\newblock Distributed deep neural networks over the cloud, the edge and end
  devices.
\newblock In \emph{IEEE International conference on distributed computing
  systems (ICDCS)}, 2017.

\bibitem[Xie et~al.(2016)Xie, Huttunen, Lin, Bhattacharyya, and
  Takala]{xie2016resource}
Xie, R., Huttunen, H., Lin, S., Bhattacharyya, S.~S., and Takala, J.
\newblock Resource-constrained implementation and optimization of a deep neural
  network for vehicle classification.
\newblock In \emph{European Signal Processing Conference}, pp.\  1862--1866,
  2016.

\bibitem[Zeng et~al.(2019)Zeng, Li, Zhou, and Chen]{zeng2019boomerang}
Zeng, L., Li, E., Zhou, Z., and Chen, X.
\newblock Boomerang: On-demand cooperative deep neural network inference for
  edge intelligence on the industrial internet of things.
\newblock \emph{IEEE Network}, 33\penalty0 (5), 2019.

\bibitem[Zhang et~al.(2021)Zhang, Xiang, Zhang, Li, Zhu, and
  Gu]{zhang2021dynamic}
Zhang, B., Xiang, T., Zhang, H., Li, T., Zhu, S., and Gu, J.
\newblock Dynamic {DNN} decomposition for lossless synergistic inference.
\newblock In \emph{IEEE International Conference on Distributed Computing
  Systems Workshops (ICDCSW)}, 2021.

\bibitem[Zhao et~al.(2018)Zhao, Barijough, and Gerstlauer]{zhao2018deepthings}
Zhao, Z., Barijough, K.~M., and Gerstlauer, A.
\newblock {DeepThings}: Distributed adaptive deep learning inference on
  resource-constrained {IoT} edge clusters.
\newblock \emph{IEEE Transactions on Computer-Aided Design of Integrated
  Circuits and Systems}, 37\penalty0 (11):\penalty0 2348--2359, 2018.

\bibitem[Zhu \& Gupta(2018)Zhu and Gupta]{zhu2017prune}
Zhu, M. and Gupta, S.
\newblock To prune, or not to prune: exploring the efficacy of pruning for
  model compression.
\newblock In \emph{International Conference on Learning Representations (ICLR)
  Workshops}, 2018.

\end{thebibliography}
\end{document}